\documentclass[conference]{IEEEconf}

\IEEEoverridecommandlockouts
\ifdefined\pdfobjcompresslevel\pdfobjcompresslevel=0\fi
\ifdefined\pdfminorversion\pdfminorversion=4\fi
\newcommand{\learningheaderfont}{\normalfont}

\usepackage{graphicx}
\usepackage{booktabs}
\usepackage{amsmath}
\usepackage{amsfonts}
\usepackage{amssymb}
\usepackage[table]{xcolor}
\usepackage{balance}
\usepackage{placeins}

\definecolor{taskWiping}{HTML}{FFF0F2}
\definecolor{taskInsertion}{HTML}{EFFBEF}
\definecolor{taskPerturbed}{HTML}{EFF3FF}
\definecolor{taskOverall}{HTML}{FFF7E8}
\definecolor{policyGray}{HTML}{F2F2F2}
\definecolor{titleRed}{HTML}{C93632}
\definecolor{titleOrange}{HTML}{D66B2C}
\definecolor{titleGold}{HTML}{AC9229}
\definecolor{titleLime}{HTML}{6C9D36}
\definecolor{titleGreen}{HTML}{278449}
\usepackage{balance}
\usepackage{placeins}

\makeatletter
\let\NAT@parse\undefined
\makeatother
\usepackage[noadjust]{cite}

\usepackage[
    colorlinks=true,
    linkcolor=blue,
    citecolor=blue,
    urlcolor=blue,
    bookmarks=true
]{hyperref}

\usepackage{caption}
\newcommand{\figref}[1]{Fig.~\ref{#1}}


\title{\LARGE \bf
\textcolor{titleRed}{P}\textcolor{titleOrange}{C}\textcolor{titleGold}{T}\textcolor{titleLime}{a}\textcolor{titleGreen}{c}: Mechanochromic Photonic-Crystal Tactile Sensor\\ for Dexterous Manipulation
}
\author{%
  Ruilin Zhang$^{1,*}$,\quad
  Xuanshuo Liu$^{1,*}$,\quad
  Junzhe Dai$^{1,2}$,\quad
  Kefei Wu$^{1}$,\quad
  Xuhao Qin$^{1}$\quad \\
  Ziyuan Jiao$^{2\dagger}$,\quad
  Wanlin Li$^{2\dagger}$,\quad
  Yurui Qu$^{1\dagger}$,\quad
  Chenxi Xiao$^{1\dagger}$%
\thanks{\textsuperscript{*} Equal contribution.}
\thanks{\textsuperscript{\dag} Corresponding authors.}
\thanks{
    $^{1}$School of Information Science and Technology,
    ShanghaiTech University, Shanghai, China.
    {\tt\footnotesize \{zhangrl2023, liuxsh2025, daijzh2023, wukf, qinxh2024, quyr, xiaochx\}@shanghaitech.edu.cn}}
\thanks{
    $^{2}$State Key Laboratory of General Artificial Intelligence,
    Beijing Institute for General Artificial Intelligence (BIGAI), Beijing, China.
    {\tt\footnotesize \{liwanlin, jiaoziyuan\}@bigai.ai}
}
\thanks{
    This work was supported by the Natural Science Foundation of Shanghai (Grant No. 25ZR1402370) and the National Natural Science Foundation of China (62405187), and was also supported by MoE Key Laboratory of Intelligent Perception and Human-Machine Collaboration (KLIP-HuMaCo) and by the State Key Laboratory of Micro-nano Engineering Science (Grant No. MES202614).
}
\thanks{
    Project webpage: \href{https://XXXX.github.io/}{https://XXXX.github.io/}
}
}

\begin{document}
\maketitle
\thispagestyle{empty}
\pagestyle{empty}

{
\begin{abstract}
\urlstyle{same}
Modern visuotactile sensors primarily sense contact by imaging the deformation of a compliant interface. Using techniques such as photometric stereo, they can recover detailed contact geometry, often represented as surface normals or depth. However, force-related information typically still needs to be inferred indirectly through calibration, geometric reconstruction, or learned mappings. Mechanochromic photonic-crystal (PC) materials offer an alternative physical encoding mechanism: contact loading changes their periodic microstructure and spectral response, directly transforming local stress into a spatially resolved structural-color field. Despite this potential, PC materials have rarely been explored as visuotactile sensors for closed-loop robot learning. We present PCTac, an open-source mechanochromic PC fingertip that uses a dense hue field as its tactile observation. Loading characterization demonstrates a repeatable relationship between the applied normal load and the observed hue response. We further integrate this representation directly into a Diffusion Policy and evaluate it on three real-robot manipulation tasks. Across 90 physical rollouts, PCTac Hue achieves 68 successes (75.6\%), compared with 60 (66.7\%) for PCTac RGB, 57 (63.3\%) for Wedge RGB, and 29 (32.2\%) for vision-only. PCTac Hue also achieves the highest observed success rate on each evaluated task, including Perturbed Test-Tube Insertion, where it achieves 16/30 successes versus 13/30 for tactile RGB baselines and 1/30 for vision-only. These results demonstrate the potential of mechanochromic structural color as a compact, load-sensitive tactile representation for robot learning and closed-loop manipulation. Project page: https://bey0ndt1me.github.io/PCTac-page/.
\end{abstract}
}

\begin{figure}[t]
    \hypertarget{image:fig:teaser}{}
    \centering
    \includegraphics[width=\columnwidth]{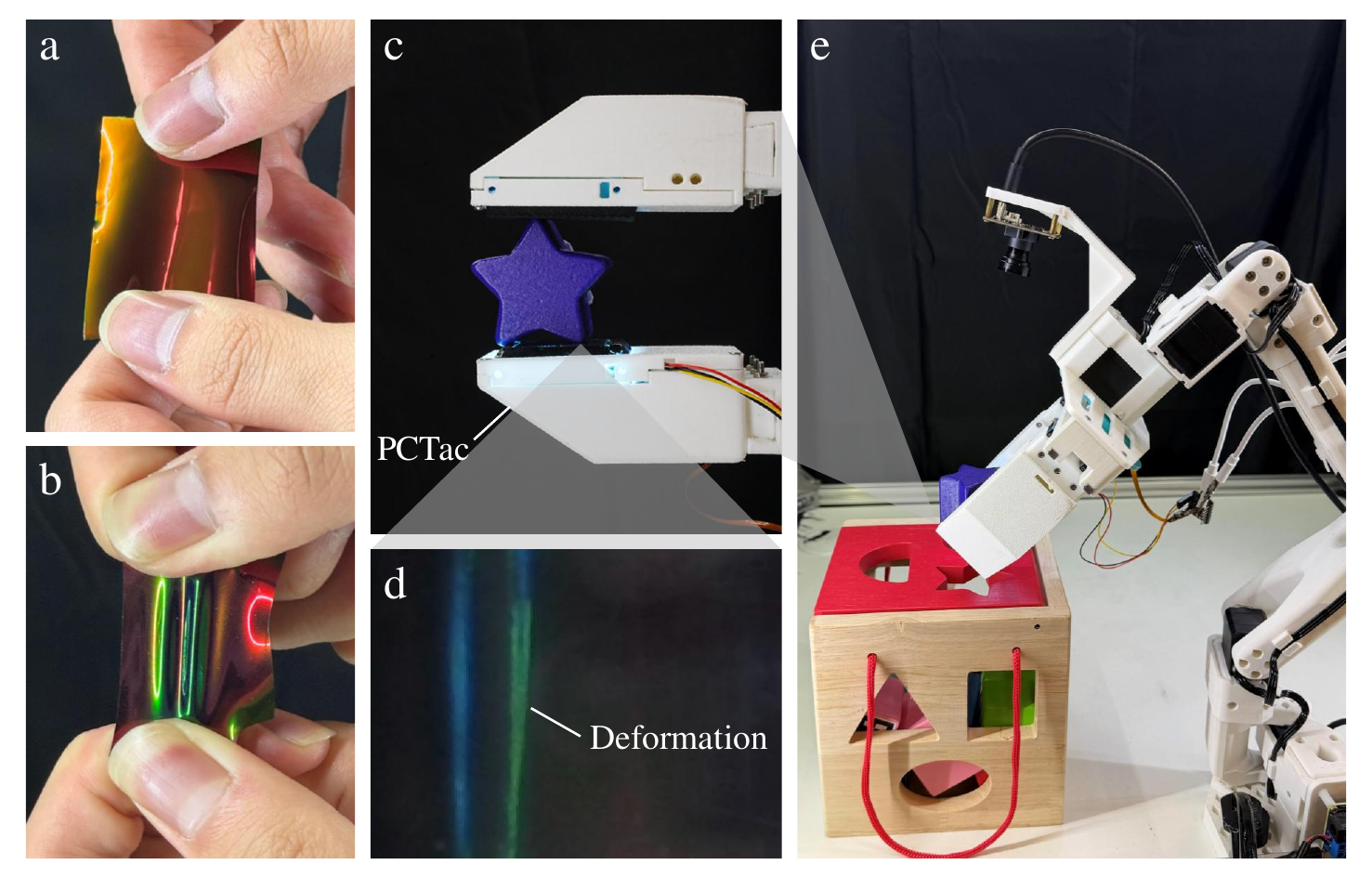}
    \vspace{-12pt}
    \caption{Overview of PCTac. (a, b) A mechanochromic PC film exhibits structural-color changes under deformation. (c) PCTac acts as the contact surface of a parallel gripper. (d) The captured color field is used  as a tactile observation without explicit surface-geometry reconstruction. (e) The sensor is integrated with the robot for contact-rich manipulation.}
    \label{fig:teaser}
    \vspace{-18pt}
\end{figure}

\section{Introduction}
{
Visuotactile sensors have emerged as a powerful sensing modality for dexterous manipulation, offering exceptionally high spatial resolution and rich multimodal contact information~\cite{Yuan2017GelSight, Wang2021Wedge, Taylor2022GelSlim, Lambeta2020DIGIT, Padmanabha2020OmniTact, Gomes2020GelTip, li2024minitac}. By imaging deformation over a compliant interface, they provide dense observations of contact geometry, local motion, and material interaction, enabling applications such as grasping, slip detection, and dexterous manipulation~\cite{Wang2021Wedge,Yuan2015Slip,Dong2019IncipientSlip,James2021Stabilization}. Their ability to capture spatially resolved contact patterns makes them particularly attractive for learning-based robot control, where tactile observations can complement vision when contact geometry and interaction dynamics are difficult to observe externally~\cite{Calandra2018Feeling,Lee2019VisionTouch}.

However, visuotactile images primarily capture local surface deformation, and extracting force-related information often requires additional calibration, geometric reconstruction, or learned mappings~\cite{Yuan2017GelSight,Taylor2022GelSlim,Sferrazza2019Optical,Ma2019DenseForce}. Moreover, their reliance on illumination and imaging optics imposes packaging constraints, making it challenging to reduce fingertip size without compromising sensing performance~\cite{Wang2021Wedge,Taylor2022GelSlim,Donlon2018GelSlim}. To bridge the gap, mechanochromic PCs offer a novel means of encoding such contact information: changes in their periodic microstructure produce corresponding changes in reflected structural color~\cite{Gao2025Instant,Sasso2026Mechanochromic}. Such material-level responses could simplify the optical system by reducing the need for engineered illumination to visualize contact-induced deformation. Although mechanochromic materials have been investigated for deformation and force sensing~\cite{Sasso2026Mechanochromic,Chen2026CLCE}, their use as tactile sensors for robotic manipulation and robot learning remains largely unexplored. In particular, whether mechanochromic structural-color changes can provide sufficiently informative and robust observations for closed-loop robotic manipulation remains an open question.

Our key insight is to treat the structural color of a mechanochromic interface as a direct robot observation rather than as an intermediate representation for reconstructing contact mechanics. Based on this idea, we develop PCTac, a mechanochromic visuotactile fingertip equipped with an embedded OV5647 camera and paired with a reference-relative hue representation. We first characterize the repeatability of its color--force response under controlled optical conditions. We then evaluate its utility in real-world tasks with Diffusion Policies~\cite{Chi2023DiffusionPolicy}, comparing task performance across visual and tactile representations.

Our experiments show that PCTac provides a repeatable relationship between recorded normal load and hue response. It can be incorporated directly into real-robot policies without explicit geometric or force reconstruction. Across three contact-rich tasks and 90 physical rollouts, PCTac Hue achieves 68/90 successes (75.6\%), compared with 60/90 (66.7\%) for PCTac RGB, 57/90 (63.3\%) for Wedge RGB, and 29/90 (32.2\%) for vision-only. This corresponds to observed gains of 8.9, 12.2, and 43.3 percentage points over these respective baselines. Task-wise, PCTac Hue achieves 24/30 successes in Whiteboard Wiping, 28/30 in Test-Tube Insertion, and 16/30 in Perturbed Test-Tube Insertion, giving the highest observed success rate among the evaluated configurations in all three tasks. Together, these results support the feasibility of using mechanochromic structural color as a direct tactile observation for closed-loop robot manipulation under the evaluated conditions.

The paper makes three main contributions.
\begin{enumerate}
    \item \textbf{PCTac fingertip.}
    We introduce PCTac, an open-source mechanochromic visuotactile fingertip that produces stress-related color signals under contact, without requiring geometric reconstruction.

    \item \textbf{Signal processing and learning integration.}
    We develop a signal-processing pipeline for extracting observations from PCTac's color response and integrating them into robot learning systems.

    \item \textbf{Validation in contact-rich manipulation.}
    We evaluate PCTac in contact-rich robotic tasks and examine the utility and limitations of mechanochromic observations for learning-based manipulation.
\end{enumerate}
\section{Related Work}

{
\subsection{Visuotactile Sensors}

Visuotactile sensors estimate contact information by imaging the deformation of a compliant interface~\cite{Johnson2009Retrographic}. GelSight-style systems recover dense contact geometry from surface appearance, enabling high-resolution measurements of contact shape and deformation~\cite{Yuan2017GelSight,Wang2021Wedge,Taylor2022GelSlim}, whereas marker-based designs track surface displacement induced by deformation~\cite{WardCherrier2018TacTip} and have been widely used for slip detection and contact-state estimation~\cite{Yuan2015Slip,Dong2019IncipientSlip,James2018TacTipSlip,James2021Stabilization}. These approaches therefore provide complementary trade-offs: image-based systems offer rich spatial observations but may require marker tracking, normal estimation, or depth reconstruction to extract task-relevant quantities, while marker-based systems provide more direct deformation measurements but with sparse spatial sampling. In particular, contact force is generally not directly observable from these visual representations and must be recovered through calibration~\cite{Yuan2017GelSight,Taylor2022GelSlim}. Recent robot-learning methods have demonstrated that tactile images or learned tactile representations can instead be directly incorporated into policies for contact-rich manipulation~\cite{Chen2023VTT,Yu2025MimicTouch,Xue2025RDP}, avoiding the need to explicitly expose such intermediate quantities to the policy. Nevertheless, the underlying observation remains primarily an image of surface appearance or deformation, leaving the downstream model to understand the physical contact state. This motivates exploring sensing representations that encode mechanical state more directly while retaining the spatial richness of image-based tactile sensing.

\subsection{Mechanochromic Structural Color}

Structural color is widespread in nature and arises from the interaction of light with periodic or nanoscale structures. For example, the iridescent colors of butterfly wings and beetle exoskeletons originate from hierarchical photonic architectures~\cite{Kinoshita2002Morpho,Parker2003Weevil}, while structural colors in bird feathers and other organisms can change with viewing or structural conditions~\cite{Osorio2002Plumage,Prum1998Feather}. Inspired by these natural photonic systems, artificial photonic structures have been developed whose optical appearance changes under mechanical deformation, enabling mechanical states to be encoded directly as visible color~\cite{Fudouzi2006PhotonicRubber,Sandt2018Optomechanical,Gao2025Instant}. Recent studies have further explored mechanochromic photonic structures for tactile sensing, demonstrating that contact-induced deformation can be directly encoded in spatially varying structural color~\cite{Sasso2026Mechanochromic,Chen2026CLCE,miller2022scalable,li2024minitac}. Table~\ref{tab:sensor_mechanisms} compares the key characteristics of mechanochromic structural-color tactile sensors with those of conventional tactile sensing approaches.

Despite these advances, existing studies have primarily focused on material fabrication and optical characterization. A systematic evaluation of structural color as a tactile observation for manipulation remains lacking, particularly regarding its utility for robot learning and control. Moreover, existing studies have not established an open and reproducible evaluation framework that connects sensing, tactile representation, robot learning, and manipulation performance. PCTac addresses this gap by treating the mechanochromic color field as the primary tactile observation and systematically evaluating it across sensor characterization, robot-learning deployment, and closed-loop manipulation.
}

\begin{table}[t]
    \definecolor{checkGreen}{HTML}{2E7D32}
    \definecolor{crossRed}{HTML}{C75450}
    \newcommand{\pctYes}{\textcolor{checkGreen}{$\checkmark$}}
    \newcommand{\pctNo}{\textcolor{crossRed}{$\times$}}
    \newcommand{\pctCite}[1]{\textsuperscript{\tiny\cite{#1}}}
    \vspace{2mm}
    \caption{\normalfont Representative visuotactile sensors.
    PCTac is open-source, and combines direct stress readout, dense sensing, and
    learning-based applications under simple (only white) illumination.}
    \label{tab:sensor_mechanisms}
    \centering
    \scriptsize
    \setlength{\tabcolsep}{1.5pt}
    \renewcommand{\arraystretch}{1.18}

    \begin{tabular}{@{}
        >{\raggedright\arraybackslash}
        p{\dimexpr0.24\columnwidth-15pt\relax}
        >{\centering\arraybackslash}p{0.23\columnwidth}
        >{\centering\arraybackslash}p{0.16\columnwidth}
        >{\centering\arraybackslash}p{0.14\columnwidth}
        >{\centering\arraybackslash}p{0.11\columnwidth}
        >{\centering\arraybackslash}p{0.13\columnwidth}
    @{}}
    \toprule
    \cellcolor{taskWiping}\textbf{Sensor}
    & \cellcolor{taskWiping}\textbf{Signal encoding}
    & \cellcolor{taskWiping}\textbf{Illumination colors}
    & \cellcolor{taskInsertion}\textbf{Direct stress readout}
    & \cellcolor{taskInsertion}\textbf{Dense field}
    & \cellcolor{taskInsertion}\textbf{Learning use} \\
    \midrule

    GelSight family$^\dagger$\pctCite{Yuan2017GelSight,Wang2021Wedge,Taylor2022GelSlim,Lambeta2020DIGIT}
    & Geometry / displacement
    & R, G, B
    & \pctNo & \pctYes & \pctYes \\

    \addlinespace[3pt]
    TacTip$^\dagger$\pctCite{WardCherrier2018TacTip,Chen2026UniForce}
    & Displacement
    & White
    & \pctNo & \pctNo & \pctYes \\

    \addlinespace[3pt]
    Finger\allowbreak Vision$\dagger$\pctCite{Yamaguchi2016FingerVision}
    & Displacement
    & Ambient
    & \pctNo & \pctNo & \pctNo \\

    \addlinespace[3pt]
    9DTact$^\dagger$\pctCite{Lin2023NineDTact}
    & Deformation
    & White
    & \pctNo & \pctYes & \pctYes \\

    \addlinespace[3pt]
    GelStereo\pctCite{Zhang2022GelStereoForce}
    & Geometry
    & White
    & \pctNo & \pctYes & \pctYes \\

    \midrule
    Sasso2026\pctCite{Sasso2026Mechanochromic}
    & Stress
    & White
    & \pctYes & \pctYes & \pctNo \\
    
    MiniTac\pctCite{li2024minitac}
    & Stress
    & White
    & \pctYes & \pctYes & \pctNo \\

    \rowcolor{taskPerturbed}
    \textbf{PCTac (ours)}$^\dagger$
    & \textbf{Stress}
    & White
    & \pctYes & \pctYes & \pctYes \\
    \bottomrule
    \end{tabular}

    \par\vspace{3pt}
    \begin{minipage}{\columnwidth}
    \scriptsize
    Direct: calibrated optical-to-stress readout without
    geometry-based inversion.
    Dense: includes reconstructed fields.
    \pctNo: not demonstrated in the cited works.
    $\dagger$: open-sourced
    \end{minipage}
    \vspace{-18pt}
\end{table}

\section{Method}
{
This section describes the sensing principle of the PC interface (Sec.~\ref{sec:pc_principle}), the mechanical and optical design of PCTac (Sec.~\ref{sec:sensor_design}), the tactile representation (Sec.~\ref{sec:tactile_processing}), and its integration into the robot-learning system (Sec.~\ref{sec:robot_system}). The overall system pipeline is shown in Fig.~\ref{fig:system_pipeline}.

\subsection{Sensing Principle of the PC Interface}
\label{sec:pc_principle}

A PC contains a periodic optical structure that selectively
reflects part of the incident white light. The reflected wavelength depends on
the spacing of this structure. For an idealized first-order reflection, the
peak wavelength can be approximated by the Bragg--Snell relation
\cite{Sasso2026Mechanochromic,Chen2026CLCE}:
\begin{equation}
    \lambda_{\mathrm{peak}}
    =
    2d
    \sqrt{
        n_{\mathrm{eff}}^2-\sin^2\theta
    },
    \label{eq:bragg_snell}
\end{equation}
where $d$ is the effective spacing between periodic lattice planes,
$n_{\mathrm{eff}}$ is the effective refractive index, and $\theta$ is the
incident angle in air. Under approximately normal incidence, this relation
reduces to $\lambda_{\mathrm{peak}}\approx2n_{\mathrm{eff}}d$.

When an object contacts PCTac, it locally deforms the PC film and the supporting
elastomer. This deformation changes the periodic structure and therefore the
reflected wavelength, producing a visible color change. Different contact
locations and deformation states produce different spatial color patterns.
The observed colors also depend on the material, illumination, and viewing
geometry~\cite{Gao2025Instant,Sasso2026Mechanochromic}.

PCTac records these color patterns with an RGB camera rather than measuring the reflection spectrum directly. We compare each tactile image with an unloaded reference image to emphasize contact-induced color changes in hue representation. The resulting color field is provided directly to the robot policy as a tactile observation. Thus, PCTac does not require explicit force estimation or surface-geometry reconstruction.
}

\subsection{Design of PCTac}
\label{sec:sensor_design}
\begin{figure}[!t]
    \hypertarget{image:fig:sensor_explosion}{}
    \centering
    \includegraphics[width=\columnwidth]{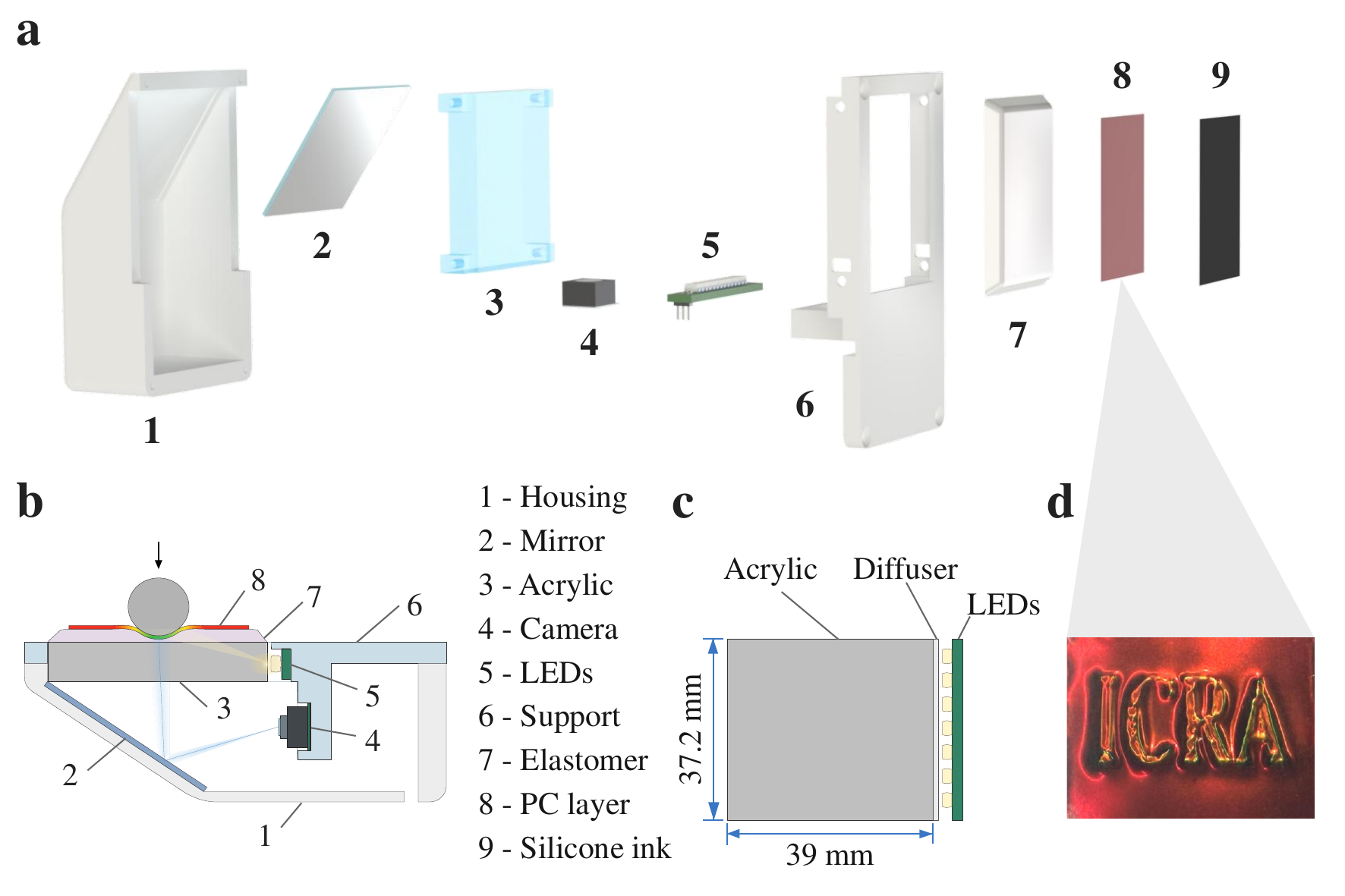}
    \caption{Mechanical and optical design of PCTac. (a) Exploded view: PLA housing (1), mirror (2), acrylic support (3), OV5647 camera (4), LEDs (5), PLA support (6), elastomer (7), PC layer (8), and silicone ink (9). (b) Cross-sectional schematic of PCTac, showing the layered contact interface and optical path. (c) One-sided illumination. (d) Image of the structural-color response to a fine contact pattern under ambient light illumination.}
    \vspace{-18pt}
    \label{fig:sensor_explosion}
\end{figure}
\begin{figure*}[!t]
    \hypertarget{image:fig:system_pipeline}{}
    \centering
    \includegraphics[width=\textwidth]{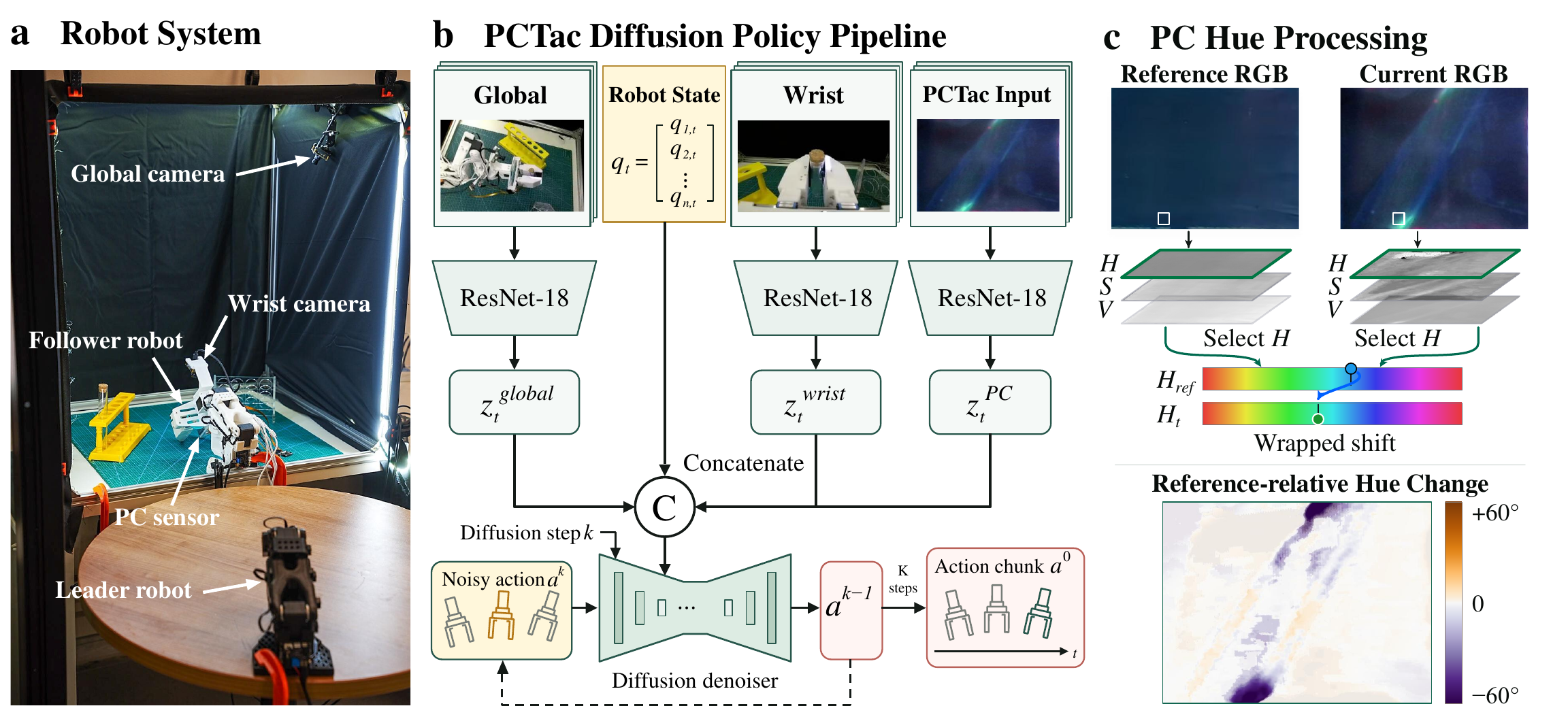}
    \caption{(a) Leader--follower data-collection setup. (b) ResNet-18 image features from PCTac, global, and wrist observations are concatenated with the robot state to condition iterative action denoising. (c) Reference and current RGB images are converted to HSV. Their hue channels are compared using a wrapped angular difference to obtain the hue-change map.}
    \label{fig:system_pipeline}
    \vspace{-18pt}
\end{figure*}
{

We next describe the design of the proposed PC-based tactile sensor, PCTac. \figref{fig:sensor_explosion} presents an exploded view of the sensor, which consists of the following components.

\noindent \textbf{(1) Sensor enclosure.} The enclosure follows the wedge-shaped design of GelSight Wedge~\cite{Wang2021Wedge}. The main body support of the sensor is printed in white PLA. The choice of white color appropriately increases the internal reflection, allowing the PC surface to receive a certain degree of vertical illumination. This increases the brightness of the PC in the unloaded state.

\noindent \textbf{(2) Camera and optical path.}
An OV5647 RGB camera is mounted near the robot flange. Because the available distance between the contact interface and the robot fingertip is limited, a mirror is used to fold the optical path and direct light from the contact surface toward the camera. This configuration increases the effective optical path length while maintaining a compact sensor form factor.

\noindent \textbf{(3) Illumination system.}
The sensor uses white LEDs to provide broadband illumination. Conventional GelSight sensors often use three-color directional illumination to reconstruct surface normals~\cite{Yuan2017GelSight,Wang2021Wedge}, whereas PCTac primarily relies on appearance changes induced at the PC contact interface. It therefore requires only spatially uniform illumination rather than separately controlled red, green, and blue lighting directions, as shown in \figref{fig:sensor_explosion}c. This simplifies the illumination hardware and reduces artifacts caused by intensity differences among independent color channels.

\noindent \textbf{(4) PC--elastomer contact interface.}
A key difference between our design and the conventional Wedge design~\cite{Wang2021Wedge} lies in the contact-interface architecture. The outermost layer is a thin coating of black silicone ink that directly contacts the manipulated object. It serves three purposes: providing a durable contact surface, increasing surface friction, and blocking ambient light from reaching the PC film. Beneath the PC film, we introduce a compliant, transparent layer of silicone elastomer (Smooth-On Solaris) between the PC and the transparent acrylic support. This compliant layer accommodates local deformation and distributes contact loads, thereby tuning the mechanical response of the PC layer. Finally, the transparent acrylic layer provides mechanical support while allowing the PC's structural-color response to be optically observed.

To fabricate the interface, we cast transparent silicone elastomer (Smooth-On Solaris) onto the acrylic support to form the elastomer layer. After curing, we bonded a $20~\mathrm{mm} \times 30~\mathrm{mm}$ PC film with a thickness of $15~\mu\mathrm{m}$ to the elastomer surface using silicone adhesive (4672 Quick-drying Glue). The elastomer thickness and stiffness jointly determine the trade-off between sensitivity and mechanical protection. We found that a thinner or stiffer elastomer generally produces a stronger structural-color response, whereas a thicker or softer elastomer provides greater mechanical protection and accommodates larger deformations. In our design, a $3~\mathrm{mm}$-thick Solaris elastomer layer (Shore A 15) was selected to balance sensitivity and durability.

{

\subsection{Tactile Signal Processing}
\label{sec:tactile_processing}

We process the camera images captured within the PC-based sensor to obtain a compact tactile representation. Although raw RGB images can be provided directly to a learning model, we found experimentally that a reference-relative circular hue representation produces more salient contact-induced features that contribute to better learning performance (See Sec.~\ref{sec:imitation_learning}). 

{
The computation for sensor signal processing proceeds as follows. Let $I_t\in\mathbb{R}^{480\times640\times3}$ denote the RGB image captured at time $t$, and let $I_0$ denote the unloaded reference image. We extract the same predefined region of interest (ROI), $\Omega$, from both images without spatial downsampling and convert the cropped images to HSV~\cite{Smith1978Color}:
\begin{equation}
    \begin{aligned}
        (h_t,s_t,v_t) &= \operatorname{HSV}(I_t|_{\Omega}), \\
        (h_0,s_0,v_0) &= \operatorname{HSV}(I_0|_{\Omega}),
    \end{aligned}
\end{equation}
where hue $h$, saturation $s$, and value $v$ are all normalized to $[0,1]$, respectively. We compute the signed hue difference
\begin{equation}
    \Delta\theta_t(u,v)
    = 2\pi\bigl(h_t(u,v)-h_0(u,v)\bigr).
\end{equation}
Encoding this angular difference by its sine and cosine avoids discontinuities at the hue boundary.

Let $m_t(u,v)$ indicate whether the current-frame saturation exceeds a predefined threshold. Using this mask, we construct the three-channel tactile encoding as follows:
\begin{equation}
    \mathbf{x}_t(u,v)=
    \begin{cases}
        \begin{bmatrix}
            Q(\sin\Delta\theta_t(u,v))\\
            Q(\cos\Delta\theta_t(u,v))\\
            255
        \end{bmatrix},
        & m_t(u,v)=1,\\[6pt]
        [127,255,0]^{\mathsf T},
        & m_t(u,v)=0,
    \end{cases}
    \label{eq:tactile_representation}
\end{equation}
where $Q(z)=\operatorname{round}[127.5(z+1)]$ maps each circular component to an 8-bit channel.

For frame-level characterization, we summarize the valid pixels using circular statistics~\cite{Fisher1993Circular}. Let $\mathcal{V}_t=\{(u,v)\in\Omega:m_t(u,v)=1\}$ denote the valid pixels at time $t$, and let $N_t=|\mathcal{V}_t|$. We compute
\begin{equation}
    \begin{aligned}
        (S_t,C_t) &= \frac{1}{N_t}\sum_{(u,v)\in\mathcal{V}_t}
        \bigl(\sin\Delta\theta_t(u,v),\cos\Delta\theta_t(u,v)\bigr),\\
        \Delta H_t &= \frac{\operatorname{atan2}(S_t,C_t)}{2\pi},
        \qquad
        R_t = \sqrt{S_t^2+C_t^2}.
    \end{aligned}
    \label{eq:force_circular_hue}
\end{equation}
Here, $\Delta H_t\in[-1/2,1/2)$ represents the circular mean hue change across valid pixels, while $R_t\in[0,1]$ measures the concentration of the hue changes.
}

\subsection{Integrated Robot Learning System}
\label{sec:robot_system}

We integrate the proposed tactile sensor into a leader--follower robotic manipulation system. The platform consists of an SO-101 leader--follower robot-arm pair, with a wrist-mounted RGB camera, a global RGB camera, and a PCTac sensor mounted on one gripper finger of the follower robot. During data collection, the operator controls the leader robot, while we record follower robot states, actions, and visual and tactile observations.

Let $\mathbf{s}_t$ denote the robot state, including the joint positions and gripper state. Let $\mathbf{v}^{w}_t$ and $\mathbf{v}^{e}_t$ denote the wrist and global camera observations, respectively, and let $\widetilde{\mathbf{x}}_t = m_t(u,v)x_t(u,v)$. The multimodal conditioning vector is
\begin{equation}
    \mathbf{c}_t
    =
    f_{\mathrm{obs}}
    \left(
        \mathbf{s}_{t-T_o+1:t},
        \mathbf{v}^{w}_{t-T_o+1:t},
        \mathbf{v}^{e}_{t-T_o+1:t},
        \widetilde{\mathbf{x}}_{t-T_o+1:t}
    \right),
    \label{eq:multimodal_condition}
\end{equation}
where $T_o$ is the observation horizon and $f_{\mathrm{obs}}$ contains the modality-specific encoders and feature-fusion layers.

We use Diffusion Policy~\cite{Chi2023DiffusionPolicy} to predict a sequence of future actions. Let
$\mathbf{A}^{0}_t=[\mathbf{a}_t,\mathbf{a}_{t+1},\ldots, \mathbf{a}_{t+T_a-1}]$ denote a clean action sequence with prediction horizon $T_a$. During training, Gaussian noise is added at diffusion step $k$~\cite{Ho2020DDPM}:
\begin{equation}
    \mathbf{A}^{k}_t
    =
    \sqrt{\bar{\alpha}_k}\,\mathbf{A}^{0}_t
    +
    \sqrt{1-\bar{\alpha}_k}\,\boldsymbol{\epsilon},
    \qquad
    \boldsymbol{\epsilon}\sim\mathcal{N}(\mathbf{0},\mathbf{I}),
    \label{eq:forward_diffusion}
\end{equation}
where $\bar{\alpha}_k$ is determined by the noise schedule. A conditional denoising network $\boldsymbol{\epsilon}_{\theta}(\mathbf{A}^{k}_t,k,\mathbf{c}_t)$ predicts the added noise and is optimized using
\begin{equation}
    \mathcal{L}_{\mathrm{DP}}(\theta)
    =
    \mathbb{E}
    \left[
        \left\|
            \boldsymbol{\epsilon}
            -
            \boldsymbol{\epsilon}_{\theta}
            \left(
                \mathbf{A}^{k}_t,k,\mathbf{c}_t
            \right)
        \right\|_2^2
    \right].
    \label{eq:diffusion_loss}
\end{equation}

For action generation during deployment, the policy initializes an action sequence with Gaussian noise and iteratively denoises it while conditioning on the current multimodal observation. The robot executes the first $T_e$ actions of the generated sequence and then replans based on the latest robot, visual, and tactile observations.

}


{
\section{Experiments}
\label{sec:experiments}

We evaluate PCTac through sensor characterization and real-robot manipulation experiments. The experiments are designed to answer three questions:

\begin{itemize}
\item \textbf{Q1: Contact response.} Does the structural-color response of PCTac vary consistently with the applied contact force?
\item \textbf{Q2: Spatial observation.} Can PCTac capture spatial patterns associated with the geometry and texture of contacting objects?
\item \textbf{Q3: Robot learning.} Can PCTac provide useful tactile observations for contact-rich imitation-learning tasks?
\end{itemize}

\subsection{Contact-Response Characterization}
\label{sec:force_characterization}

{
We characterize the hue--force relationship by recording the reference normal load and tactile images during indentation at the center and four peripheral contact locations. A multi-axis force sensor (Photon Finger, Haptron Scientific Inc.) measures the applied load, while the hue is computed as the circular mean over a fixed rectangular image region and normalized to $[0,1]$. We define the compression-positive normal load as $F_n=-F_z$, where $F_z$ denotes the normal-force channel of the reference sensor. \figref{fig:force_response}b, c summarizes 22 normal-loading trials, with the loading and unloading phases shown separately to characterize the optical response and hysteresis. The color beneath the loading curve represents the measured hue. The hue changes substantially with increasing load, with hysteresis observed between loading and unloading due to elastomer recovery. At the indented location, the sensor saturates at approximately 1.5~N. This saturation is specific to the compact spherical indenter with a diameter of 5~mm used in the experiment; higher loads can be measured when the contact area is larger.
}

\begin{figure}[t]
    \hypertarget{image:fig:force_response}{}
    \centering
    \includegraphics[width=\columnwidth]{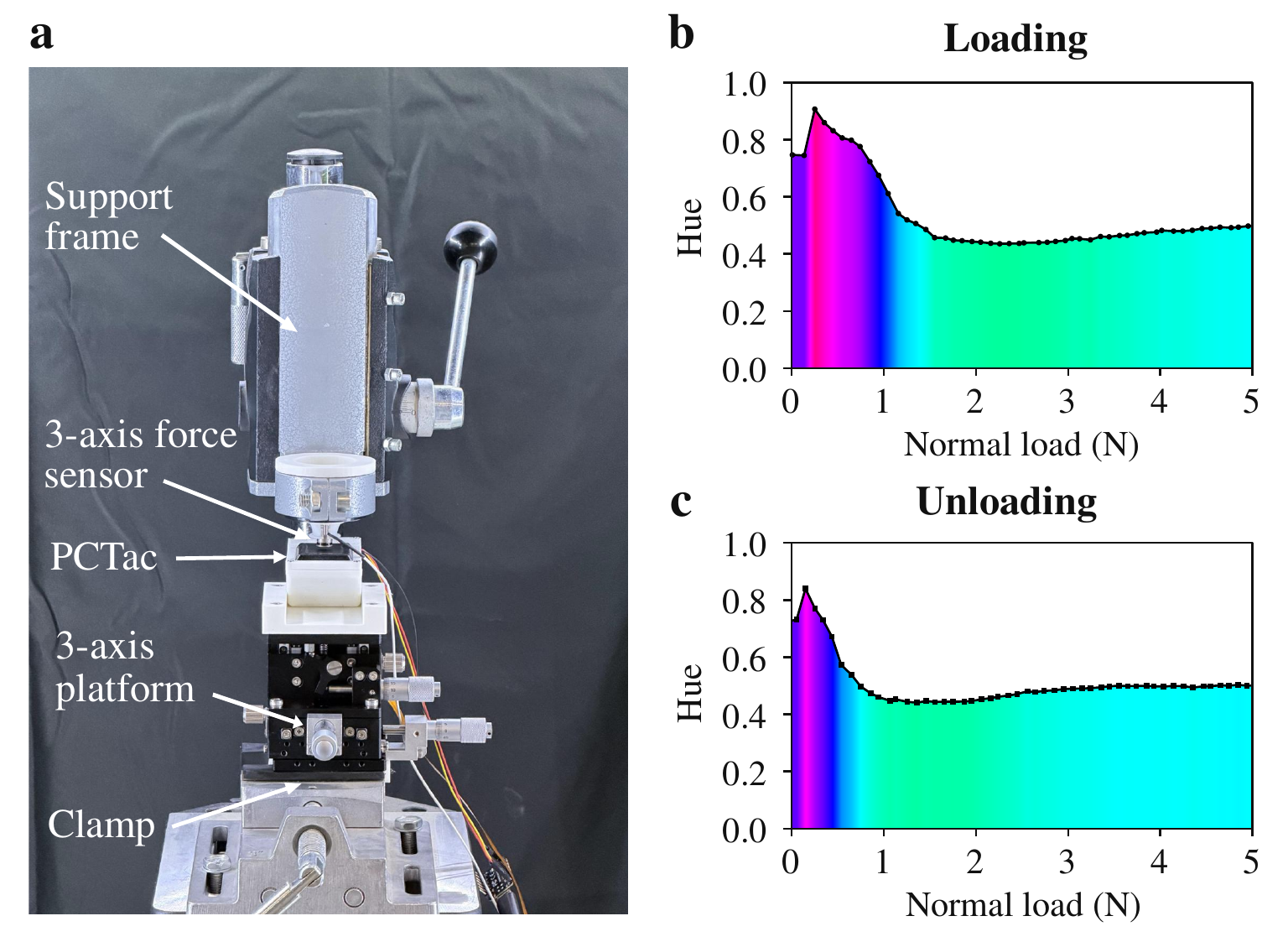}
    \caption{Hue--force characterization. (a) Experimental apparatus. (b, c) Circular mean hue versus normal load during loading and unloading.}
    \label{fig:force_response}
    \vspace{-18pt}
\end{figure}

}

{
\subsection{Spatial Observation of Contacting Objects}
\label{sec:object_observation}

\begin{figure*}[t]
    \hypertarget{image:fig:object_samples}{}
    \centering
    \includegraphics[width=\textwidth]{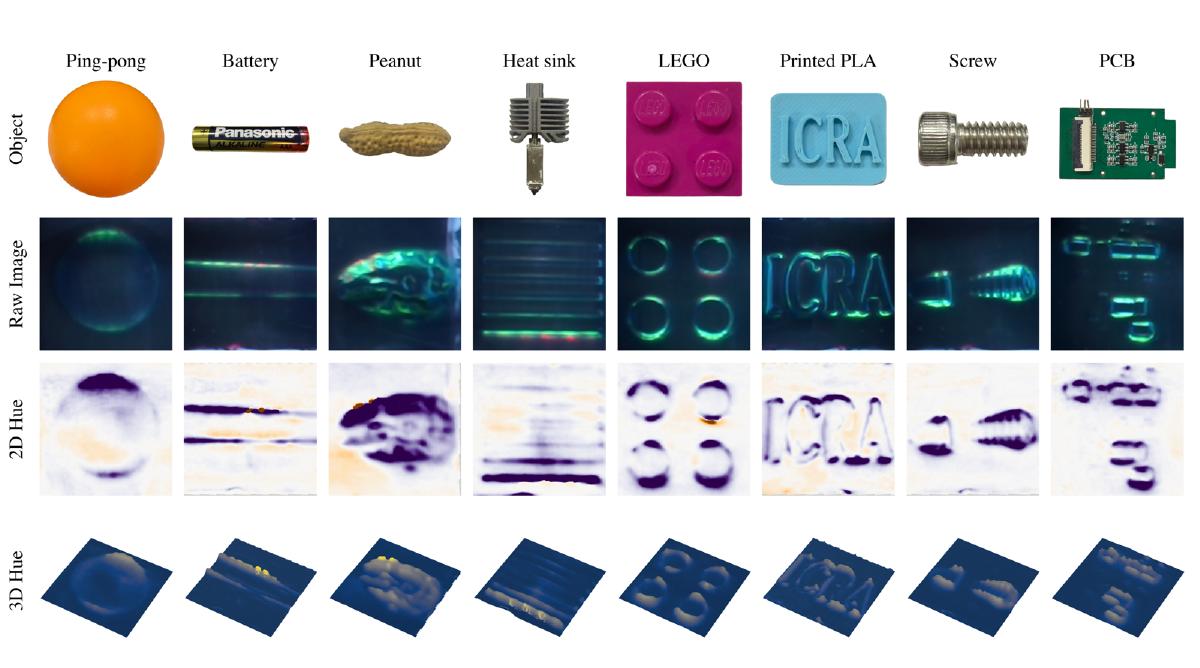}
    \caption{PCTac observations for objects with different geometries and textures. Each group shows the object photograph, raw tactile RGB image, circular hue-difference field, and a three-dimensional visualization of the hue-difference field.}
    \label{fig:object_samples}
    \vspace{-18pt}
\end{figure*}

We next examine the ability of PCTac to resolve contact regions and surface features. Eight objects with diverse geometries and surface structures are used: a Ping-pong ball, a battery, a peanut, a heat sink, a LEGO brick, a printed PLA sample, a screw, and a printed circuit board (PCB). These samples encompass diverse contact geometries and surface structures, including flat and curved surfaces, repeated ridges, raised patterns, and fine threads.

Each object is pressed against the sensor using a clamping device under a fixed normal load of 10~N. For each object, we show a photograph, a raw tactile RGB image, a circular hue-difference field, and a three-dimensional rendering of the same field. The vertical coordinate in the three-dimensional rendering represents hue change rather than physical surface height.

\figref{fig:object_samples} presents representative tactile observations. Raised features on the LEGO brick and PCB produce distinct boundaries in the PCTac responses, while repeated structures on the heat sink produce clearly resolved ridge patterns. Fine features, such as the printed PLA texture and screw threads, remain discernible in the hue representation. In contrast, larger convex objects, such as the Ping-pong ball and battery, produce smoother and more spatially extended responses due to their broader contact regions. The peanut exhibits less pronounced recessed surface features because the sensor has limited conform to recessed regions. Improving the representation of such geometries is left for future work.


\subsection{Imitation Learning}
\label{sec:imitation_learning}

We further evaluate whether PCTac provides task-relevant observations for robot imitation learning through three contact-rich manipulation tasks.

\noindent\textbf{(1) Tasks.}
The evaluation includes three tasks with different contact requirements. We collected 60 trials of teleoperation demonstrations per task for policy training. Each policy is evaluated in 30 physical rollouts per task, for 90 rollouts in total. 
\begin{itemize}
    \item \textit{T1: Whiteboard Wiping} requires the robot to grasp an eraser, wipe a randomly marked region of a whiteboard while maintaining stable contact, and place the eraser into a bowl. To facilitate grasping, we designed a rigid holder for the back of the eraser. Due to occlusion, the external cameras were not able to clearly observe the eraser–gripper and eraser–whiteboard contact regions. Hence, tactile feedback complements visual observation during wiping. 
    \item \textit{T2: Test-Tube Insertion} requires the robot to grasp a test tube, adjust its in-hand orientation to align it with the target rack, and insert it into the designated hole within the rollout horizon. This task evaluates precise orientation control and contact-aware alignment.
    \item \textit{T3: Perturbed Test-Tube Insertion} additionally evaluates recovery from an unexpected change in the grasped object's orientation. After grasping and before final insertion, a human operator rotates the tube away from the target insertion direction, and the robot must restore the tube's alignment with the target hole before completing the insertion.
\end{itemize}

Representative successful trajectories for the three tasks are shown in
\figref{fig:rollouts}.

\begin{figure}[!t]
    \hypertarget{image:fig:rollouts}{}
    \centering
    \includegraphics[width=\columnwidth]{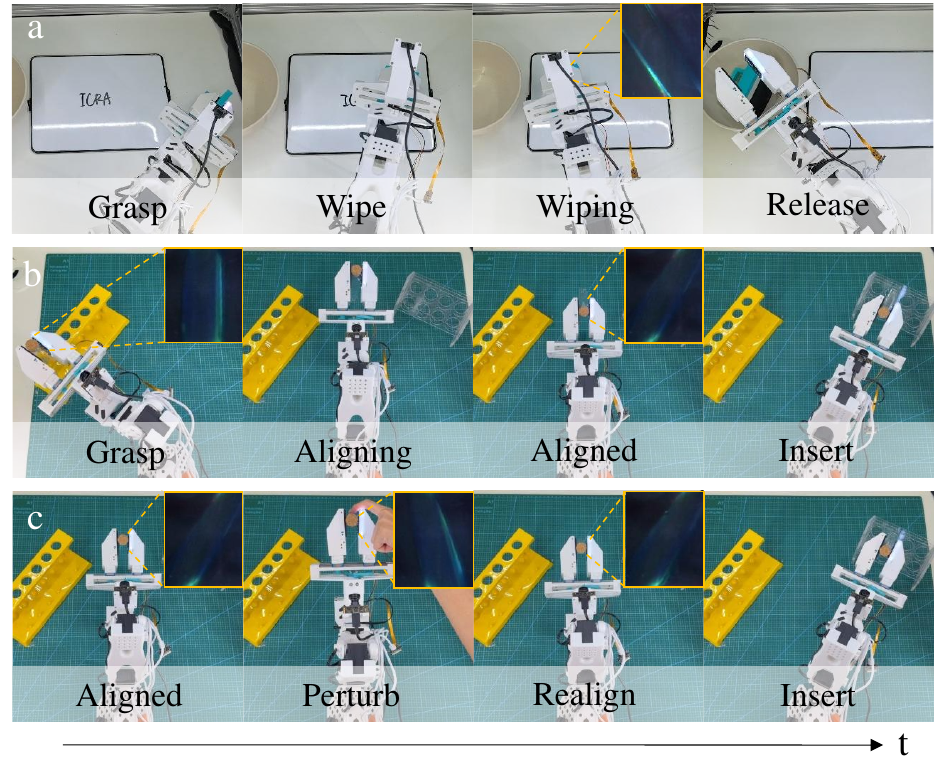}
    \caption{Representative successful trajectories with raw tactile images. (a) Whiteboard Wiping. (b) Test-Tube Insertion. (c) Perturbed Test-Tube Insertion.}
    \label{fig:rollouts}
    \vspace{-22pt}
\end{figure}

\noindent\textbf{(2) Policy Configurations.}
All policies use Diffusion Policy~\cite{Chi2023DiffusionPolicy} and receive the same types of inputs: robot state, wrist-camera images, and global-camera images. The configurations differ in their tactile observations.
The Vision-only baseline receives no tactile observations.
The Wedge RGB baseline uses raw RGB images from our implementation of GelSight Wedge.
Two PCTac configurations use either raw RGB images (PCTac RGB) or the reference-relative circular hue representation defined in Eq.~\eqref{eq:tactile_representation} (PCTac Hue).

All configurations use ResNet-18 encoders~\cite{He2016ResNet} for visual and tactile observations. The observation history, action horizon, policy architecture, training schedule, checkpoint selection rule, and action-chunk execution procedure are kept consistent. The vision-only configuration excludes tactile input to assess the contribution of tactile sensing, whereas the tactile configurations compare different sensing hardware and representations.

\begin{table*}[t]
    \vspace{2mm}
    \caption{\normalfont Imitation-learning performance across three contact-rich tasks: whiteboard wiping, test-tube insertion, and perturbed insertion. Bold numbers denote the best performance in each column.}
    \label{tab:imitation_learning_comparison}
    \centering
    \small
    \setlength{\tabcolsep}{7pt}
    \renewcommand{\arraystretch}{1.18}
    \begin{tabular}{@{}>{\columncolor{policyGray}[0pt][\tabcolsep]}l>{\columncolor{policyGray}}l>{\columncolor{taskWiping}}c>{\columncolor{taskInsertion}}c>{\columncolor{taskPerturbed}}c>{\columncolor{taskOverall}[\tabcolsep][0pt]}c@{}}
        \toprule
        {\learningheaderfont Policy}
        & {\learningheaderfont Tactile representation}
        & {\learningheaderfont\shortstack{Whiteboard\\Wiping\\($n=30$)}}
        & {\learningheaderfont\shortstack{Test-Tube\\Insertion\\($n=30$)}}
        & {\learningheaderfont\shortstack{Perturbed\\Insertion\\($n=30$)}}
        & {\learningheaderfont\shortstack{Overall\\($n=90$)}} \\
        \midrule
        Vision-only
        & None
        & 19/30 (63.3\%)
        & 9/30 (30.0\%)
        & 1/30 (3.3\%)
        & 29/90 (32.2\%) \\

        Wedge RGB
        & Raw RGB
        & 23/30 (76.7\%)
        & 21/30 (70.0\%)
        & 13/30 (43.3\%)
        & 57/90 (63.3\%) \\

        PCTac RGB
        & Raw RGB
        & 22/30 (73.3\%)
        & 25/30 (83.3\%)
        & 13/30 (43.3\%)
        & 60/90 (66.7\%) \\

        PCTac Hue
        & Circular hue difference
        & \textbf{24/30 (80.0\%)}
        & \textbf{28/30 (93.3\%)}
        & \textbf{16/30 (53.3\%)}
        & \textbf{68/90 (75.6\%)} \\
        \bottomrule
    \end{tabular}
    \vspace{-18pt}
\end{table*}

\noindent\textbf{(3) Results.}
The experimental results are summarized in Table~\ref{tab:imitation_learning_comparison}. Across all 90 rollouts, PCTac Hue achieves an overall success rate of 75.6\%, compared with 32.2\% for vision-only and 63.3\% for Wedge RGB. PCTac Hue thus improves the overall success rate by 43.3 and 12.2 percentage points over vision-only and Wedge RGB, respectively. 

In Whiteboard Wiping, PCTac Hue achieves the highest success rate of 80.0\%, surpassing Wedge RGB by 3.3 percentage points. The largest performance gap occurs in Test-Tube Insertion, where PCTac Hue outperforms Wedge RGB by 23.3 percentage points. This advantage is maintained in Perturbed Test-Tube Insertion, the most challenging task, where PCTac Hue achieves the highest success rate of 53.3\%. This result suggests that PCTac Hue may help the policy handle the evaluated orientation perturbations. We also find that the choice of tactile representation affects policy performance: across all three tasks, PCTac Hue consistently outperforms PCTac RGB. Overall, PCTac Hue achieves the highest success rate across all three tasks, demonstrating the effectiveness of the proposed structural-color-based tactile representation for contact-rich manipulation.

The main failure modes across the three tasks are misgrasping and misinsertion (Fig.~\ref{fig:failure_samples}). These failures typically occur before meaningful physical contact with the target and are therefore potentially attributable to visual perception and policy generalization rather than tactile perception. Failures can also occur during tactile interaction. For example, in Whiteboard Wiping, occasional failures include residual marks and missing the bowl when releasing the eraser. In Test-Tube Insertion, the robot occasionally becomes stuck and fails to proceed with the task. We believe these failures are largely caused by inputs that fall outside the training distribution. Increasing the diversity and coverage of the training data could therefore improve policy generalization and reduce such failures.
}


\subsection{Discussion}
\label{sec:experiment_discussion}

PCTac Hue directly maps contact-induced structural-color changes to tactile observations, without explicitly reconstructing contact geometry. This differs from conventional visuotactile representations based on estimated surface normals, depth, or marker motion~\cite{Yuan2017GelSight,Wang2021Wedge,Taylor2022GelSlim}. Such tactile signals can be used directly as inputs to learning-based policies, where explicit geometric reconstruction may not be necessary. Moreover, the structural-color sensing principle simplifies the illumination design, reducing optical complexity while retaining informative tactile responses. These properties make PCTac Hue a compact and practical tactile representation for robot manipulation.

Our system has two main limitations. First, the current elastomer is less sensitive to concave features than to raised ones. Raised patterns produced more pronounced and localized color changes, whereas concave features often resulted in weaker spatial responses. We attribute this to the limited conformability of the current elastomer, which could be improved through material improvements. Second, because the photonic crystal primarily responds to local stress, different combinations of normal and shear deformation can produce similar color patterns. Future work will explore elastomer designs with improved sensitivity to concave features and methods to better distinguish such ambiguous responses.

}

\begin{figure}[t]
    \hypertarget{image:fig:failure_samples}{}
    \centering
    \includegraphics[width=\columnwidth]{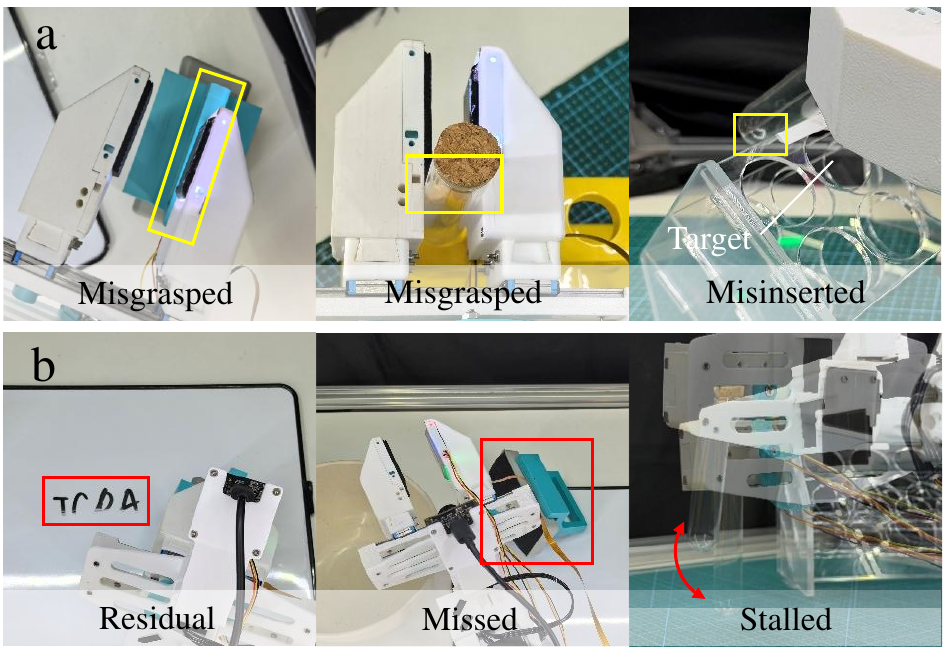}
    \caption{Representative failures across the three tasks. (a) Misgrasping and misinsertion potentially associated with visual perception and policy generalization. (b) Occasional failures such as residual marks, a missed drop, and a stalled manipulation.}
    \label{fig:failure_samples}
    \vspace{-18pt}
\end{figure}

\FloatBarrier
\section{Conclusion}

We presented PCTac, a mechanochromic PC fingertip that directly converts contact-induced structural-color changes into tactile observations for robot manipulation. By exploiting structural color directly, PCTac enables a simplified optical design while preserving useful contact information. We further introduced a reference-relative circular hue representation that captures spatial and load-dependent color changes, providing an effective tactile input for robot policy learning without requiring explicit force or geometric reconstruction. In robot experiments with Diffusion Policy, PCTac Hue achieved an overall success rate of 75.6\% across Whiteboard Wiping, Test-Tube Insertion, and Perturbed Insertion, compared with 66.7\% for PCTac RGB and 32.2\% for vision-only. These results demonstrate that structural color can provide a compact and effective tactile representation for contact-rich robot learning. Future work will improve the ability to distinguish tactile responses to normal and shear loading and evaluate PCTac across a broader range of contact conditions and manipulation tasks.

\bibliographystyle{IEEEtran}
\bibliography{references}
\end{document}